\pdfoutput=1

\documentclass[11pt]{article}

\usepackage[]{latex/acl}

\usepackage{times}
\usepackage{latexsym}
\usepackage{graphicx}
\usepackage{algorithm}
\usepackage{amsmath} 
\usepackage{algorithmic}
\usepackage{amsfonts}
\usepackage{bm} 
\usepackage{multirow}
\usepackage{multicol}
\usepackage{natbib}
\usepackage{booktabs}
\usepackage{adjustbox}

\usepackage[T1]{fontenc}

\usepackage[utf8]{inputenc}

\usepackage{microtype}

\usepackage{inconsolata}

\title{Parallel Decoding via Hidden Transfer for\\ Lossless Large Language Model Acceleration}

\author{ \textbf{Pengfei Wu\textsuperscript{1,2}\thanks{\enspace Equal contribution.}, Jiahao Liu\textsuperscript{3}\footnotemark[1], Zhuocheng Gong\textsuperscript{1}, Qifan Wang\textsuperscript{4}, Jinpeng Li\textsuperscript{1}} \\ 
\textbf{Jingang Wang\textsuperscript{3}, Xunliang Cai\textsuperscript{3}, Dongyan Zhao\textsuperscript{1,2,5,6}\thanks{\enspace Corresponding authors: Dongyan Zhao (zhaody@pku.edu.cn).}}\\
\textsuperscript{1}Wangxuan Institute of Computer Technology, Peking University \\
\textsuperscript{2}Center for Data Science, AAIS, Peking University; \textsuperscript{3}Meituan; \textsuperscript{4}Meta AI \\
\textsuperscript{5}National Key Laboratory of General Artificial Intelligence; \textsuperscript{6}BIGAI, Beijing, China \\
\texttt{\{pengfeiwu1999,lijinpeng\}@stu.pku.edu.cn} \\
\texttt{\{liujiahao12,wangjingang02,caixunliang\}@meituan.com} \\
\texttt{wqfcr@fb.com}, \texttt{\{zhaody,gzhch\}@pku.edu.cn}}

\begin{document}
\maketitle
\begin{abstract}
Large language models (LLMs) have recently shown remarkable performance across a wide range of tasks. However, the substantial number of parameters in LLMs contributes to significant latency during model inference. This is particularly evident when utilizing autoregressive decoding methods, which generate one token in a single forward process, thereby not fully capitalizing on the parallel computing capabilities of GPUs.
In this paper, we propose a novel parallel decoding approach, namely \textit{hidden transfer}, which decodes multiple successive tokens simultaneously in a single forward pass. The idea is to transfer the intermediate hidden states of the previous context to the \textit{pseudo} hidden states of the future tokens to be generated, and then the pseudo hidden states will pass the following transformer layers thereby assimilating more semantic information and achieving superior predictive accuracy of the future tokens.

Besides, we use the novel tree attention mechanism to simultaneously generate and verify multiple candidates of output sequences, which ensure the lossless generation and further improves the generation efficiency of our method.
Experiments demonstrate the effectiveness of our method. We conduct a lot of analytic experiments to prove our motivation. In terms of acceleration metrics, we outperform all the single-model acceleration techniques, including Medusa and Self-Speculative decoding.

\begin{figure}[!t]
  \centering
  \includegraphics[width=1.\linewidth]{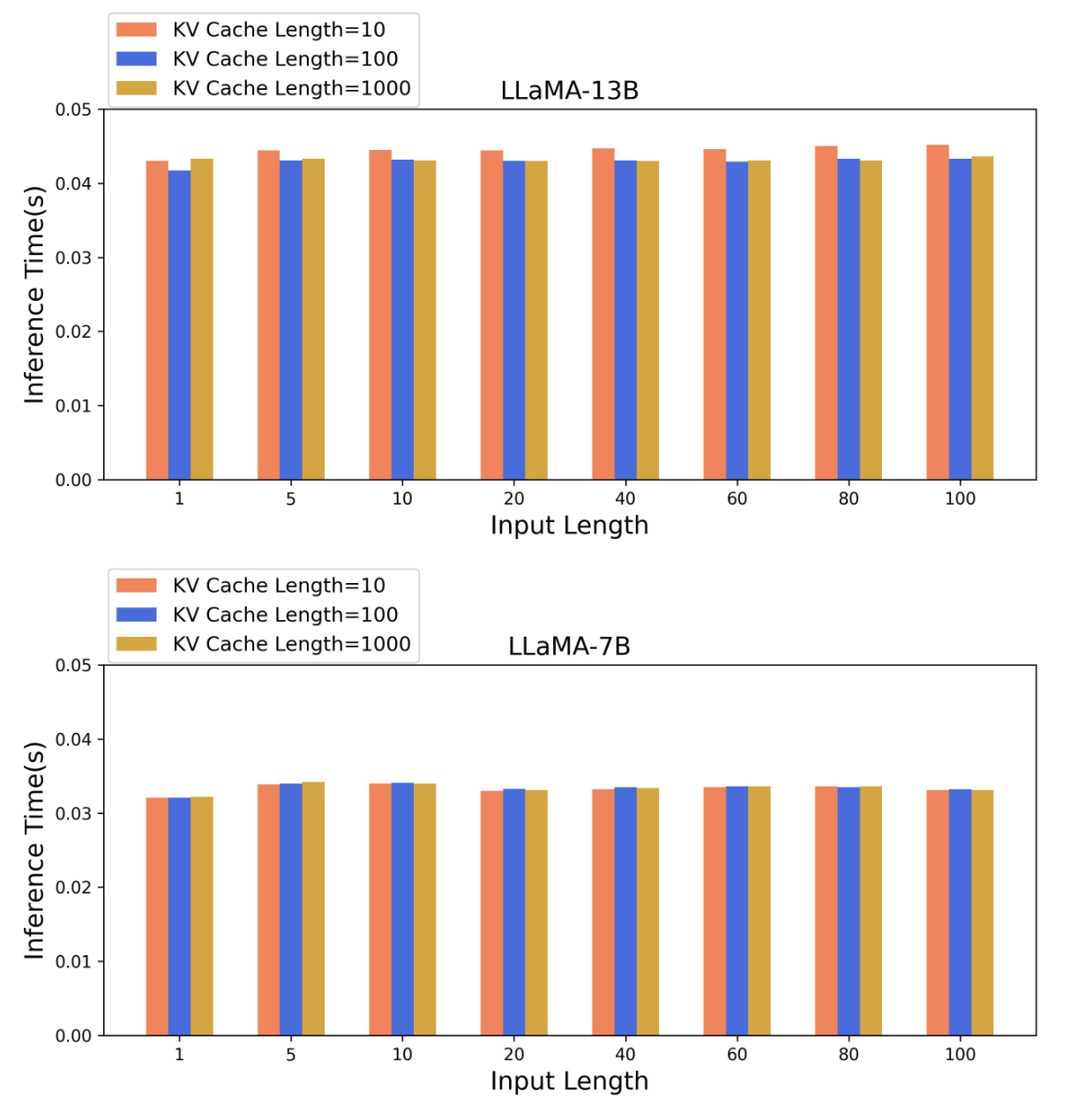}
  \caption{A single forward time consumption for various LLMs with different size under different KV cache length and different number of input tokens, each data is the average of randomly 100 samples. The result shows that under the setting of KV cache, increasing the length of input sequence in a certain range will not increase the time of forward propagation, and then prove that the traditional autoregressive decoding has a waste in GPU utilization efficiency}
  \label{fig:kv_cache_inference_time}
\end{figure}

\end{abstract}

\section{Introduction}
Recent developments in Transformer-based large language models~\citep{vaswani2017attention, radford2018improving, radford2019language, brown2020language, zeng2022glm, zhang2022opt, touvron2023llama, touvron2023llama-2, roziere2023code} have demonstrated remarkable performance across a broad spectrum of tasks. However, these models grapple with excessive inference latency due to the inherently serial process of generating one token per forward pass, the inference process is typically memory bandwidth-bound, which means most of the time model inference is spent loading billion parameters from memory rather than computing, resulting in the waste of the parallel computational power of GPUs~\cite{shazeer2019fast}. In our experiments, We find that due to the parallelism of GPU computing, the time for multiple tokens to propagate in parallel is nearly the same as the time for one token to propagate(as shown in Table~\ref{fig:kv_cache_inference_time}). Therefore, the low efficiency of the inference stage becomes the biggest bottleneck to broaden the application scenarios of LLMs.

To address this bottleneck, contemporary work proposes speculative decoding~\citep{leviathan2023fast, chen2023accelerating, zhang2023draft}, which utilizes a small language model to draft a few tokens ahead. then the LLM verifies the drafted tokens and accepts the correct ones. While this acceleration technique achieves promising performance, it has its limitations: speculative decoding requires another model to do the draft thing. 
It is inconvenient to cooperate with an extra model in some scenarios as it requires more sophisticated scheduling and the draft model might consume extra GPU and memory resources.
Thus, some researchers have been investigating single-model acceleration. That is,  to speed up the inference of LLMs without auxiliary models. Self-speculative decoding~\cite{zhang2023draft} and Medusa~\cite{medusa} are typical methods within this line of research.
Medusa predicts not only the next token within a single forward propagation but also a few tokens ahead of the next token. These extra tokens are predicted based on the last hidden states of the input tokens through the trainable Medusa heads.


Our method aligns with the line of single-model acceleration. We design a novel method called \textbf{Hidden Transfer}, to predict the \textit{pseudo} hidden states of future tokens in the intermediate layers. We use a trainable linear projection to transfer the hidden states of input tokens to the pseudo hidden states of future tokens in a certain intermediate layer, and the synthesized pseudo hidden states pass the subsequent layers and interact with hidden states of the whole sequence as normal, in the last layer, we use the original lm-head and decode the draft tokens of the future positions.
In this way, we can predict more than merely the next token but also a few tokens ahead in a single forward propagation. In the training stage we employ KL-divergence as the supervised signal, which minimizes the distribution between tokens predicted by the pseudo hidden states and the real ones. 
In addition to the novel design of Hidden Transfer, we also use the tree attention mechanism~\citep{medusa, miao2023specinfer, spector2023accelerating} to simultaneously perform the token prediction and token verification to ensure the lossless generation of our method.

The motivation for hidden transfer is that the synthesized pseudo hidden states will interact with themselves and previous hidden states of the context during the forward propagation in which gain more semantic information to boost the success rate of predicting future tokens.
Our experiments show that this motivation is fulfilled, and our method can achieve the best draft token prediction accuracy and inference acceleration ratio compared with other methods under a single-model setting.

Our key contributions are:
(1) To our best knowledge, we are the first to study the prediction of pseudo hidden states of the future tokens in LLMs, our experiments prove that intermediate hidden states could be predicted directly and refined in the forward propagation.
(2) We propose Hidden Transfer, a novel single-model lossless acceleration method for improving the inference efficiency of LLMs. Our method predicts multiple draft tokens with synthetic pseudo hidden states.
(3) We conduct various experiments to prove the effectiveness of our method, including some analytic experiments to prove our motivation.

\section{Related Work}
To solve the problem of inference latency in LLMs, the existing works can be divided into the following two categories: we call the first category \textbf{Model Compression}, including model distillation~\cite{sanh2019distilbert}, model pruning~\citep{frantar2023sparsegpt, wang2021spatten} and model quantization~\cite{liu2023llm}, aiming to replace the original large language model with a small model which have the similar but not identical outputs in certain field. We refer the second category of methods as \textbf{Speculative Decoding}, the key idea of is to reduce the number of forward propagation of the LLMs under the condition that the generated results remain unchanged.


\subsection{Model Compression}
There are plenty of works focus on the model lightweight, including model quantization~\citep{han2015deep, zhao2019improving, jacob2018quantization, yao2022zeroquant, frantar2022gptq, liu2023llm}, knowledge distillation~\citep{hinton2015distilling, cho2019efficacy,hsieh2023distilling}, model pruning~\citep{xia2023sheared, guo2023compresso, chen2023lorashear} and model sparsification~\citep{hoefler2021sparsity, liu2023deja}. Model quantization is to convert the model parameters to floating-point numbers with lower precision or integers; Model pruning and sparsification removes redundant components in the LLMs; Knowledge distillation works by transferring knowledge from a teacher model to a student model (small model). These model compression methods have a wide range of applications, but they do not guarantee that the output is strictly consistent with the original LLMs.

\subsection{Speculative Decoding}
This series of methods aim to quickly generate some draft tokens and use the LLMs to verify them in parallel to keep lossless generation theoretically. We can divide this class of methods into two types based on the number of deployed models, single model and multiple models. 
The single models' approach is represented by Medusa~\cite{medusa}, and Self-speculative decoding method~\cite{zhang2023draft}, Medusa and train extra multiple heads to predict subsequent tokens based on the last hidden state of the input tokens after a single forward propagation. Self-speculative methods use a subset of intermediate layers of the whole LLM as the draft model to generate draft tokens
The multiple models' approach is represented by traditional speculative decoding~\citep{leviathan2023fast, chen2023accelerating}, which uses a small language models~(SLMs) as the draft model to generate draft tokens, the LLMs verify the tokens in parallel.

\section{Methodology}
In this section, we first define the problem formulation, including the overview of traditional autoregressive decoding algorithm and parallel decoding algorithm, then we provide a detailed description of the training and testing process of our hidden transfer method.   
\begin{figure*}[t]
  \centering
  \includegraphics[width=1.\linewidth]{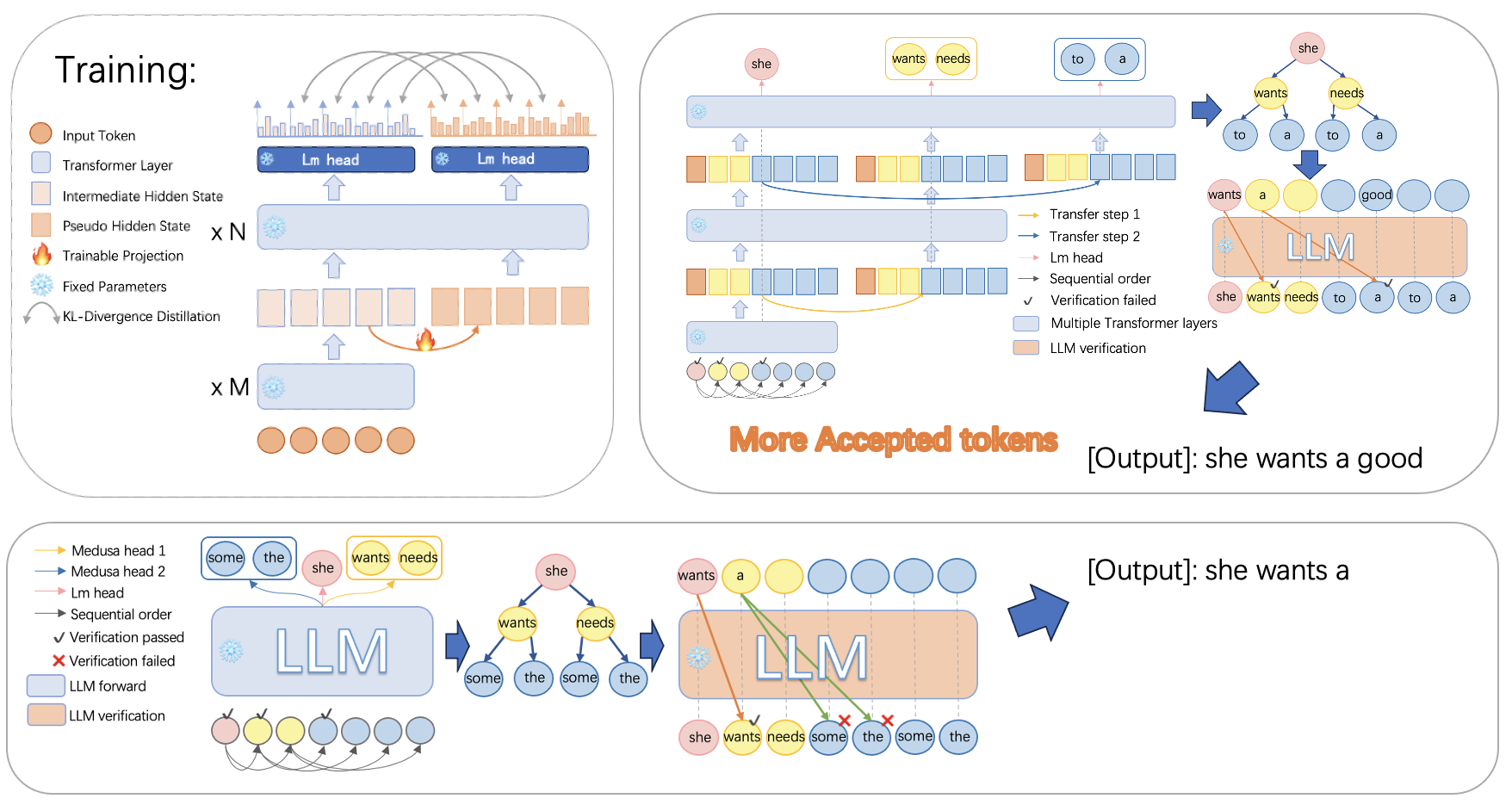}
  \caption{Overview of our method. The upper-left is the training process of hidden transfer, $N$ and $M$ represent the numbers of transformer layers. The upper-right and the bottom of the figure are the inference process of Hidden transfer and Medusa respectively, assuming that the generation of both methods starts from the same context and their inputs are the candidate token sequences generated in the last round, Medusa and Hidden transfer both verify the candidate token sequences to find the last accepted token position(i.e the fourth token of the input) and generate the next token and new draft tokens at the same time(for simplicity we only consider 2 transfer steps/medusa heads), the next token and draft tokens construct to a tree structure and are then flatted into a sequence to be verified with tree attention, after the verification stage, result shows Hidden transfer has more prediction accuracy and more draft tokens accepted}
  \label{fig:main}
\end{figure*}

\subsection{Problem Formulation}
Transformer-based auto-regressive language models aim to construct the the ${(n+1)}_{th}$ token's distribution given the prefix $n$ tokens, denoted as $P(x_{n+1} | x_1, x_2, \ldots, x_n)$
These models are capable of processing entire sequences in parallel during the training stage. However, during inference stage, the generation process becomes serial naturally. This is due to the requirement of the preceding $n$ tokens' semantic information to predict the probability distribution of the ${(n+1)}_{th}$ token. Traditional autoregressive generation techniques, whereby a single token is produced per model forward process, fail to capitalize on the parallel processing capabilities of GPUs, consequently impacting the response efficiency of various AI systems. The essence of parallel decoding algorithms, exemplified by speculative decoding, lies in their ambition to increase the expected number of tokens generated in one single forward propagation of the LLMs while maintain the generation consistency. Thus, the optimization goal of this class of methods can be written to find a maximum positive integer $k$ that satisfies the following conditions:

{
\small
\[
\tilde{P}(X_{n+k+1}\ldots X_{n+1}| X_{\leq n}) = P(X_{n+k+1}\ldots X_{n+1}|X_{\leq n})
\]
}

\noindent Where $P$ and $\tilde{P}$ represent the token distribution given by the original language model and parallel decoding algorithm respectively. For most parallel decoding algorithms, $\tilde{P}$ is obtained by a draft and verify process, tree attention is wildly adopted to verify multiple candidate sequences simultaneous, so a lot of works focus on how to generate better candidates in the draft stage, the process can be described as the following formula:

{
\[
X_{n+k+1}, \ldots, X_{n+1} = M(X_{\leq n})
\]
}

In some speculative decoding algorithms, $M$ represents small language models, or part of the original LLMs~\cite{zhang2023draft}, in the block-wise series of methods (represented by Medusa~\cite{medusa}), $M$ can be viewed as the extra heads in the last layer of LLMs, In our method, $M$ can be viewed as the hidden transfer linear projection in some intermediate layers, in the following section, we concisely introduce our method, including the training and inference stages.

\subsection{Hidden Transfer}
The core idea of our hidden transfer is to predict the future $k+1$ tokens(including $k$ draft tokens and the next token generated correctly by language model) by one step of LLM forward propagation without the deployment of SLMs, existing works under this setting either use earlying exiting method to directly predict the token distribution~\cite{bae2023fast} or train extra $k$ lm-heads to predict the $k$ draft tokens in the last layer, we believe that the first method will lose a lot of information at higher layers, assuming that we use $X_{n}$ to represent the $n_{th}$ token of the input sequence, and $h_{n}^{j}$ represents the $j_{th}$ layer's hidden state of the $n_{th}$ token, the first method trains an early lm-head to predict $X_{n+1}$ in advance, but once the $X_{n+1}$ is predicted, it should be used as the input in the next round of forward propagation, the previous forward propagation will stop in the middle layer, as a result the higher layers' hidden state of $X_{n}$ will be lost(i.e. $h_{n}^{j+1}$, $h_{n}^{j+2}$ $\ldots$), although there're some works claim that they can simply use copy mechanism to simulate the higher layers' hidden state~\cite{elbayad2019depth}, but other experiments still show that the copy mechanism performs badly in some cases~\cite{bae2023fast}. The second method which use extra lm-heads to predict the future $k$ draft tokens is very simple and effective, but we believe this method lacks the interaction of the $k$ draft tokens and the previous tokens because the draft tokens do not interact with the previous tokens through the attention mechanism, for example it only use the last hidden states of $X_{n}$ to predict the $X_{n+2}$ without using the $X_{n+1}$'s information, but some times the $X_{n+2}$ depends on the $X_{n+1}$, so our method choose to train multiple transfer functions (simple linear projections) to \textbf{predict} the future $k$ draft token's hidden states in some intermediate layers by mapping the $h_{n}^{j}$ to $h_{n+1}^{j}$ ... $h_{n+k}^{j}$, so we can continue the forward propagation with $n+k$ intermediate hidden states, During the following transformer layers, the last $k$ hidden states will pass the original lm-head to predict the token distribution normally, the whole training and inference process compared with Hidden transfer and Medusa is shown in Figure\ref{fig:main}. 

\subsection{Training stage}
At the training stage, it is required to train multiple linear projections in multiple fixed layers, where the locations of the corresponding transfer layers and the number of transfer step are considered as hyperparameters. The number of transfer step is equal to the number of pseudo hidden states predicted in a single forward process for one token; hence, we denote this number as $k$. Because we train $k$ linear projections separately so we need to conduct the training process $k$ times(we train one linear projection for one transfer step in a certain layer). For simplicity we discuss the training process of the $i_{th}$ transfer step, we denote the index of transfer layer for step $i$ as $t_{i}$. so we can use $W_{t_{i}}^{i} \in \mathbb{R}^{d \times d}$ ($1 \leq i \leq k$) to denote the trainable linear projection for the $i_{th}$ step($d$ represent the hidden dimension of the LLMs). Assume we have original token sequences $X_{1}$, $X_{2}$, $...$, $X_{n}$, we first do the forward process to the $t_{i}$ layer to get their corresponding hidden state $h_{1}^{t_{i}}$, $h_{2}^{t_{i}}$, $...$, $h_{n}^{t_{i}}$, then we transfer all of the $n$ hidden states into their corresponding pseudo hidden states, which can be formulated as below:

{
\[
\widetilde{h}_{n}^{t_{i}}, \widetilde{h}_{n-1}^{t_{i}}, \ldots, \widetilde{h}_{1}^{t_{i}} = W_{t_{i}}^{i} \cdot (h_{n}^{t_{i}}, h_{n-1}^{t_{i}}, ...,h_{1}^{t_{i}}) 
\]
}

After transferring the $h_{j}^{t_{i}}$ into $\widetilde{h}_{j}^{t_{i}}$ ($1 \leq j \leq n$), we concat the original hidden states and the pseudo hidden states into a new sequence (ie.$h_{1}^{t_{i}},..., h_{n-1}^{t_{i}},h_{n}^{t_{i}}, \widetilde{h}_{1}^{t_{i}}, \ldots, \widetilde{h}_{n-1}^{t_{i}}, \widetilde{h}_{n}^{t_{i}}$). It's easy to show that the $\widetilde{h}_{j}^{t_{i}}$ is the pseudo hidden state of $h_{j+1}^{i}$, so in the self-attention layers it can only view the hidden states from $h_{1}^{t_{i}}$ to $h_{j+i-1}^{t_{i}}$ and itself in the sequence, we design attention mask to achieve the goal. In order to ensure the consistency between the training stage and inference stage, we set the position id $j+i$ to $\widetilde{h}_{j}^{t_{i}}$ to construct the position embedding. 


We then continue to forward the new sequence, and get the final representations of the last layer (i.e., $h_{1}^{l}, \ldots, h_{n-1}^{l}, h_{n}^{l}, \widetilde{h}_{1}^{l}, \ldots, \widetilde{h}_{n-1}^{l}, \widetilde{h}_{n}^{l}$, where $l$ denotes the number of layers of the LLMs). We then pass the hidden states sequence through the original lm-head and finally get the token distribution of each position ($P_{1}^{l}, \ldots, P_{n-1}^{l}, P_{n}^{l}, \widetilde{P}_{1}^{l}, \ldots, \widetilde{P}_{n-1}^{l}, \widetilde{P}_{n}^{l}$). We use the KL-divergence between token distributions given by the pseudo hidden states and the original hidden states as the supervised signal. The loss can be formulated as below:

{\small
\[
\small
Loss_{distll} = \sum_{q=1}^{n-i-1} KL-divergence(\widetilde{P}_{n+q}^{l}, {P}_{q+i}^{l})
\]

}











\subsection{Inference stage}

In the inference stage, the process is to first generate some sequence candidates and then verify them, the purpose of the verification stage is to ensure the tokens consistency with the auto-regressive decoding, We construct multiple sequence candidates into a tree structure by merging their common ancestors, then we flat the tree into a sequence and construct attention mask correctly to keep their orders, finally we send the whole sequence into the LLMs to verify the candidates and generate the new candidates at the same time. Figure\ref{fig:main} shows the inference stage of both Hidden transfer and Medusa.

\subsubsection{Tree attention}
When predicting the draft tokens for the following several steps, it becomes clear that the draft tokens belonging to different steps form a tree structure according to their order. The tree attention mechanism transforms this hierarchical tree into a linear sequence while preserving the original positional indices of each token. Moreover, it employs a specialized attention mask to ensure that a token only attends to its ancestors within the tree structure, thereby upholding the causal language model's properties. During inference, a pre-defined tree structure and parser facilitate the rapid transformation of token candidates between the tree and sequence representations, obviating the need for additional computations.

\begin{table*}
\begin{adjustbox}{width=0.9\width,center}
\begin{tabular}{lcccc}
\toprule
\textbf{Model} & \textbf{Decoding Algorithm}& \textbf{XSum} & \textbf{Gsm8k}\\
\midrule
\multirow{4}{*}{LLaMA-2-Chat-13B} & {Auto-regressive} & 1.000$\times$  & 1.000$\times$ \\ 
& {Self-speculative}~\cite{zhang2023draft}  & 1.241$\times$ & 1.216$\times$\\
& {Medusa}~\cite{medusa} & 1.325$\times$  & 1.976$\times$ & \\
& \textbf{Ours} & \textbf{1.532}$\times$  & \textbf{2.275}$\times$  \\
\midrule
\multirow{4}{*}{Vicuna-13B}  & {Auto-regressive} & 1.000$\times$ & 1.000$\times$  \\ 
                              & {Self-speculative}~\cite{zhang2023draft} & 1.125$\times$& 1.118$\times$   \\
                           & {Medusa}~\cite{medusa} & 1.247$\times$  & 1.869$\times$\\
                           & {\textbf{Ours}}  & \textbf{1.419}$\times$ & \textbf{2.150}$\times$ \\
\midrule
\multirow{3}{*}{LlaMA-2-Chat-7B}  & {Auto-regressive} & 1.000$\times$  & 1.000$\times$ \\ 
                           & {Medusa}~\cite{medusa} & 1.465$\times$  & 1.762$\times$\\
                           & {\textbf{Ours}}  & \textbf{1.816}$\times$  &\textbf{2.135}$\times$ \\
\midrule
\multirow{3}{*}{Vicuna-7B}  & {Auto-regressive} & 1.000$\times$& 1.000$\times$ \\ 
                           & {Medusa}~\cite{medusa} & 1.388$\times$  &  1.732$\times$\\
                           & {\textbf{Ours}}  & \textbf{1.786}$\times$  & \textbf{2.219}$\times$ \\
\bottomrule
\end{tabular}
\end{adjustbox}
\caption{The acceleration ratio for both forward times and end-to-end time}
\label{tab:main experiment}
\end{table*}

\section{Experiment}
\subsection{Setup}
We evaluate the our method on two different series of models with different size, including LLaMA-2-CHAT-13B~\cite{touvron2023llama-2}, LLaMA-2-CHAT-7B~\cite{touvron2023llama-2} and Vicuna-13B~\cite{vicuna2023}, Vicuna-7B~\cite{vicuna2023}. We use greedy sampling strategy for the LLMs and the tokens generated using our method are identical to those generated by standard auto-regressive decoding theoretically. We divide the experiments into main experiments subsection, analytical and ablation study. The main experiments are conducted on all models and the analytical and ablation study only conducted on 7B model for simplicity.

In the main experiments subsection, we compare our end-to-end time acceleration ratio with Medusa~\cite{medusa} and Self-speculative decoding~\cite{zhang2023draft} to show the effectiveness of our method(more details in Appendix), Because the self-speculative decoding method~\cite{zhang2023draft} need to carefully select the transformer layers skipped for each model, and it doesn't offer the initial skip layers of 7B model in its open source code for their optimizer algorithm, so we only compare with them on LLaMA-2-Chat-13B and vicuna-13B, we use the acceleration ratio reported in their paper of LLaMA-2-Chat-13B for the Xsum dataset, and run its open source code to evaluate their effectiveness on the Gsm8K dataset and vicuna-13B. we use the same Tree structure and predict the next three draft tokens (exclude the next token generated by the original lm-head) for Medusa and our method, and we do hidden transfer for our method on the 25 30 35 layers respectively.

Analytical experiments subsection and ablation study aim to verify our motivation, So we first compare the draft tokens prediction accuracy between our method and two baselines (e.g. Medusa heads and Early existing on different intermediate layers), result shows that we have the best prediction accuracy for the future draft tokens in a single forward propagation; To verify our motivation, we also analyze how the hidden states similarity between the pseudo hidden states predicted and the original hidden states changing along with the forward propagation and prove the \textbf{refinement} of transformer layers, finally we train multiple transfer projections on different layers for different transfer steps to explore how to select the transfer layers. Finally we also compare the transfer prediction accuracy of the second transfer step between the setting of first pseudo hidden states masked or not in the ablation study

All experiments are conducted on a single NVIDIA A100-80GB GPU and all implementations are based on PyTorch using HuggingFace’s architecture~\cite{wolf2020transformers, lhoest2021datasets}
\begin{figure*}[!t]
  \centering
  \includegraphics[width=1.\linewidth]{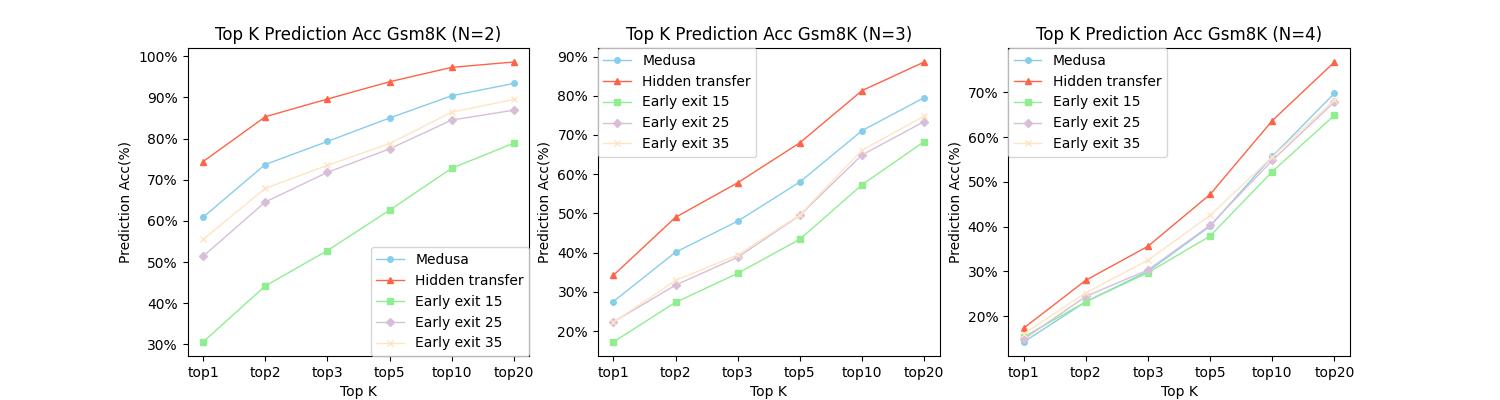}
  \caption{TopK tokens' prediction accuracy using three prediction methods on LLaMA-2-Chat-13B model including directly train different lm-heads on some intermediate layers (denoted as Early exit in the figure), Medusa method and our hidden transfer method (We transfer the pseudo intermediate hidden states of the next 3 tokens on the $25_{th}$, $30_{th}$ and $35_{th}$ layers respectively), The $N$ in the figure is the prediction step (N=2 means we predict the first draft token). It's clear that our method achieve the best prediction accuracy}
  \label{fig:token_acc}
\end{figure*}

\subsection{Datasets}
We use ShareGPT dataset as our training dataset for all models, and we use the test split of Extreme Sum marization (XSum)~\cite{narayan2018don}, Gsm8k~\cite{cobbe2021training} as our test dataset. ShareGPT is a multi-round conversations dataset comprises nearly 70,000 samples, We train one epoch for all the models. XSum~\cite{narayan2018don} is a dataset for evaluation of abstract single-document summary systems, it's test split has 11,334 samples. we only sample 1000 sentences followed~\cite{zhang2023draft}. The Gsm8k dataset~\cite{cobbe2021training} encompasses a collection of 8,500 linguistically varied, high-quality math word problems for grade school students, all of which were meticulously crafted by human authors, we use its whole test split with 1000 samples. All the two datasets are evaluated under 1-shot setting followed~\cite{zhang2023draft}.

\subsection{Main Results}

Table~\ref{tab:main experiment} shows that the acceleration ratio of our method is significantly better than other baselines in end-to-end time for all the test dataset(more details in appendix), it has at most \textbf{1.28x} acceleration ratio compared with Medusa, which is similar to our method. Using hidden transfer to predict the pseudo hidden states in the intermediate layer gain more benefit in the overall performance than using Medusa head to predict the token distribution directly, this improvement is more obvious in 7B models, and we find that the acceleration ratio on Gsm8k is higher because the answer in Gsm8k is more logical and predictable with more mathematical symbols.

\subsection{Analytical Study}
In this subsection, we conduct some analytical experiments to further verify the effectiveness and motivation of our method, the key idea of our method is to predict draft tokens more correctly in a single forward propagation by predicting the pseudo hidden states in intermediate layers. So we first compare our method with Medusa and early exiting to show that we have better draft tokens prediction accuracy; then we analyze how the hidden states change during the forward propagation to verify the refinement we proposed. We also verify the draft tokens prediction dependency and show how to select the transfer layers.

\paragraph{Draft tokens prediction accuracy}
In a single forward propagation (given the token sequence $X_1$, ... $X_{n}$ and model need to predict the future $K$ draft tokens $\widetilde{X}_{n+2}$, ... $\widetilde{X}_{n+k+1}$), so we first compare the draft tokens' prediction accuracy on three different methods: early exiting in the intermediate layers, using medusa heads and our hidden transfer method. Early exiting method trains independent lm-heads in several intermediate transformer layers to directly predict the draft tokens(for example train a lm-head and use it to map $h_{n}^{j}$ to $\widetilde{X}_{n+2}$),  We use the LLaMA-2-Chat-13B as the base model for this experiment, three different methods are trained on ShareGPT dataset for one epoch and we set $K$ as 3. We random sample 100 sequences from the test split of XSum and Gsm8K dataset respectively, for each sequence, we random split 50 points at its output part, and for each split-point, we use the token sequence before as the prompt input and predict $K$ draft tokens using different methods and compare with the tokens generated greedily by the original model in the following $K$ steps.(We choose five random seeds and average the results to better eliminate random errors) Figure~\ref{fig:token_acc} shows that our hidden transfer method achieve the best prediction accuracy among them.

\begin{figure}[!t]
  \centering
  \includegraphics[width=1.\linewidth]{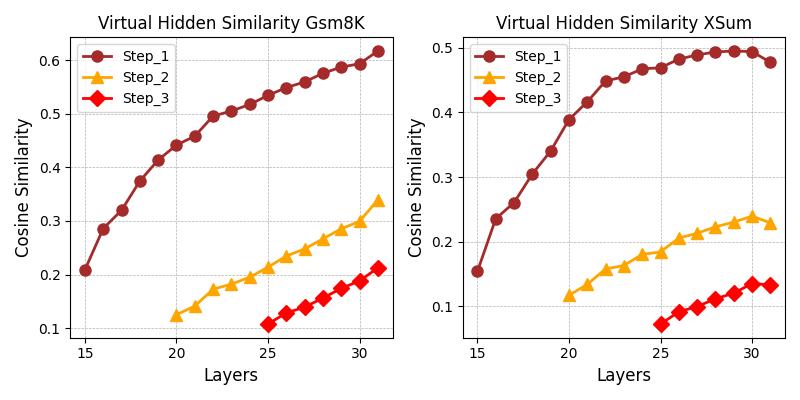}
  \caption{Hidden states similarity between the virtual hidden states predicted and the original hidden states.}
  \label{fig:hidden_similarity}
\end{figure}

\paragraph{Pseudo hidden states refinement}
 we conduct another experiment to prove that the forward propagation in transformer layers can \textbf{refine} the predicted hidden states by given more semantic information using self-attention mechanism. We compare the cosine similarity of the pseudo hidden states predicted with the \textbf{original} hidden states, and trace how the cosine similarity changes with the forward process (for example, we denote the prompt sequence as $X_1$, ... $X_{n}$ and we calculate the cosine similarity between pseudo hidden states $\widetilde{h}_{n+1}^{t}$ and real $h_{n+1}^{t}$, $h_{n+1}^{t}$ is the hidden state of $X_{n+1}$ on the $t_{th}$ layer and $X_{n+1}$ is the greedy decoding output token given the $n$ prefix context). We random sample 100 sequences for two datasets and random split 50 times for each sequence as well. Figure~\ref{fig:hidden_similarity} shows the cosine similarity get closer along with the forward process, which proves that with the forward process, the hidden states can be refined by the transformer layers.

\begin{figure}[!t]
  \centering
  \includegraphics[width=1.\linewidth]{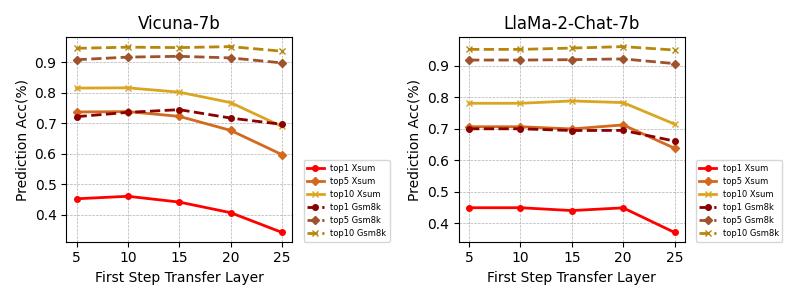}
  \caption{The first transfer step prediction accuracy on different layers for Vicuna-7b and LlaMa-2-Chat-7b. TopK means the topk tokens predicted by the transfer step include}
  \label{fig:different layers acc step1}
\end{figure}

\begin{figure}[!t]
  \centering
  \includegraphics[width=1.\linewidth]{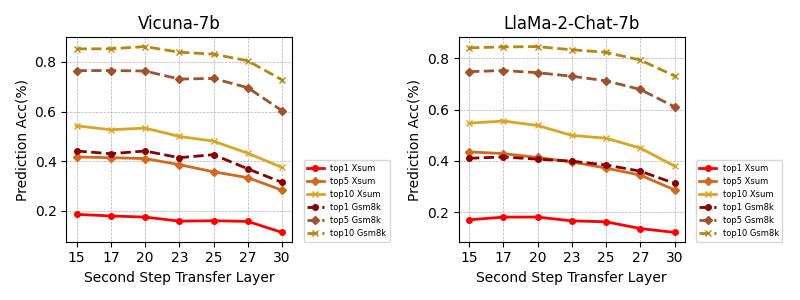}
  \caption{The second transfer step prediction accuracy on different layers for Vicuna-7b and LlaMa-2-Chat-7b with the fixed transfer step 15. TopK means the topk tokens predicted by the transfer step include}
  \label{fig:different layers acc step2}
\end{figure}

\paragraph{How to choose transfer layers}
In the inference stage of our method, we need to choose which layer to transfer and generate the pseudo hidden states, there's a trade-off between accuracy and efficient: if we transfer on the lower layers, the pseudo hidden states will pass more subsequent layers and take more computing resource but gain more semantic information by interacting with the hidden states of context; if we transfer on higher layers, the pseudo hidden states will take less computing resource with less semantic information. So we train the first and second transfer step on different layers of a fixed LLM to study the impact of the transfer layer selection. We choose Vicuna-7B and LlaMa-2-7b-chat model. For the first transfer step, we train different transfer structure on the $5_{th}$, $10_{th}$, $15_{th}$, $20_{th}$, $25_{th}$ layers seperately, and report the token prediction accuracy in figure~\ref{fig:different layers acc step1}, we found that from the lower layer to the middle layer, the prediction accuracy is basically unchanged, and from the middle layer to the high layer, the prediction accuracy drops rapidly, which proves that the middle layer to do tranfer is an optimal choice for both accuracy and computational efficiency, so we choose the $15_{th}$ layer to conduct the first transfer step. After fixed the first transfer layer, we also train different transfer structures on the layers higher than it as our second transfer layer. Figure~\ref{fig:different layers acc step2}
shows that the second transfer step have the same rule, starting from the $15_{th}$ layer, the accuracy rate remains unchanged within a range, and then rapidly decline, so we choose $20_{th}$ layer to conduct the second transfer step.

\subsection{Ablation study}
It's clear that Medusa~\cite{medusa} predict the draft tokens in parallel which means the generation between different draft tokens is independent. Our motivation is that serialized generation of draft tokens will gain better performance and we use experiment to prove it. We compare the prediction accuracy of the second transfer step under two setting: \textbf{Masked} and \textbf{No masked}, \textbf{Masked} means the second pseudo hidden state can not see the first pseudo hidden state in the forward process, and \textbf{No masked} is the normal self-attention mechanism. Figure~\ref{fig:hidden_similarity} shows that in all the models and datasets, \textbf{Masked} performs worse which prove that the semantic information of the first draft token is important to the generation of the second draft token.
\begin{figure}[!t]
  \centering
  \includegraphics[width=1.\linewidth]{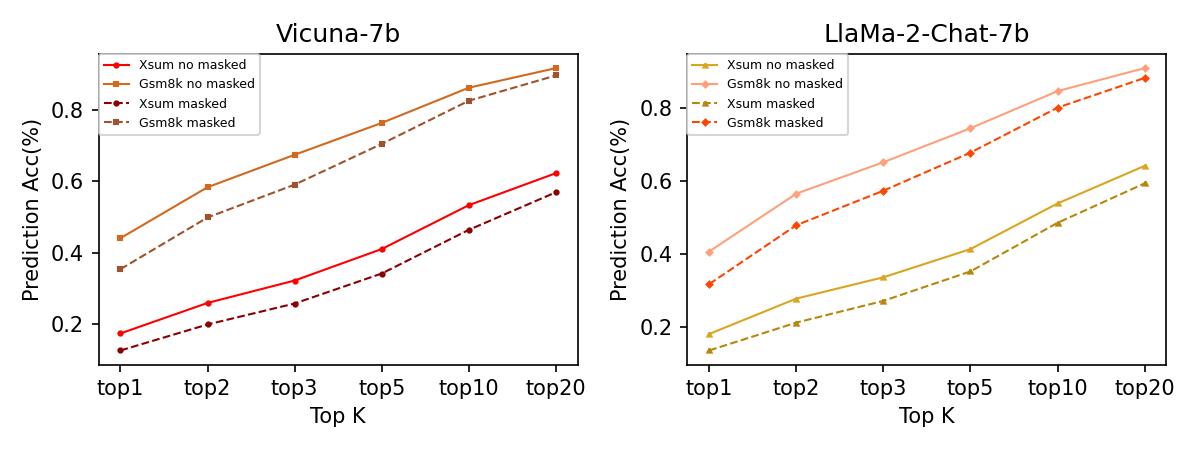}
  \caption{The second transfer step prediction accuracy under masked and no masked setting, we transfer the hidden states on $15_{th}$ and $20_{th}$ respectively}
  \label{fig:transfer_step2_need_attention.jpg}
\end{figure}

\section{Conclusion}
In this paper, we introduce a novel parallel decoding approach, named hidden transfer, designed for accelerating inference in large language models. By training a linear transformation projection in the intermediate layers, our model is capable of predicting the pseudo hidden states of multiple subsequent tokens in a single forward propagation. These predicted hidden states obtain additional semantic information through subsequent transformer layers, resulting in enhanced prediction accuracy. Through analytical experiments, we have proved that the hidden states predicted by the intermediary layers are progressively refined, gaining increased semantic information in the subsequent transformer layers by interacting with context. Our experiments demonstrate that our method outperforms existing approaches in terms of predictive precision within a single forward iteration and also achieves substantial gains in generation velocity.

\section*{Limitation}
In the verification stage of our method, we simply utilized vanilla tree attention without specific optimization. However, the structural choices of tree attention significantly impact the generation speed. Therefore, future work will focus on optimizing it. Furthermore, we aim to conduct a more extensive set of analytical experiments to elucidate the underlying mechanisms of hidden transfer better and design improved training methodologies to enhance the quality of draft token generation. Our approach necessitates the expansion of the input sequence during both training and inference, which may lead to an increase in computational resource requirements. This issue will be addressed in the future research.

\bibliography{reference}

\end{document}